%% file: main.tex
\documentclass{article}

\usepackage{arxiv}

\usepackage[utf8]{inputenc} 
\usepackage[T1]{fontenc}    
\usepackage{hyperref}       
\usepackage{url}            
\usepackage{booktabs}       
\usepackage{amsfonts}       
\usepackage{nicefrac}       
\usepackage{microtype}      
\usepackage{lipsum}
\usepackage{moreverb,url}
\usepackage{graphicx}
\usepackage{amssymb,amsmath}

\usepackage{comment}
\usepackage{xcolor}
\usepackage{comment}
\usepackage{xspace}
\usepackage{amsthm,mathrsfs,dsfont} 
\usepackage{times}
\usepackage{latexsym}
\usepackage{multicol}
\usepackage{helvet} 
\usepackage{courier}  
\usepackage{subcaption}

\newcommand{\PB}[1]{\textcolor{red}{PB: #1}}

\newcommand{\tlDCRNN}{\texttt{TL-DCRNN}\xspace}
\newcommand{\ssDCRNN}{\texttt{SS-DCRNN}\xspace}

\newcommand{\trainss}{\texttt{train-src}\xspace}
\newcommand{\valss}{\texttt{val-src}\xspace}
\newcommand{\testss}{\texttt{test-src}\xspace}
\newcommand{\traints}{\texttt{train-tgt}\xspace}
\newcommand{\valts}{\texttt{val-tgt}\xspace}
\newcommand{\testts}{\texttt{test-tgt}\xspace}

\usepackage{algorithm} 
\usepackage{algpseudocode} 
\algrenewcommand\algorithmicrequire{\textbf{Input:}}
\algrenewcommand\algorithmicensure{\textbf{Output:}}

\title{Transfer Learning with Graph Neural Networks for \\Short-Term Highway Traffic Forecasting}

\author{
  Tanwi Mallick \\
  Mathematics and Computer Science Division\\
  Argonne National Laboratory, Lemont, IL \\
  \texttt{tmallick@anl.gov} \\
   \And
 Prasanna Balaprakash \\
  Mathematics and Computer Science Division \& 
  Argonne Leadership Computing Facility \\
  Argonne National Laboratory, Lemont, IL \\
  \texttt{pbalapra@anl.gov} \\
  \And
 Eric Rask\\
  Energy Systems Division\\
  Argonne National Laboratory, Lemont, IL \\
  \texttt{erask@anl.gov} \\
  \And
  Jane Macfarlane\\
  Sustainable Energy Systems Group \\
  Lawrence Berkeley National Laboratory, Berkeley, CA \\
  \texttt{jfmacfarlane@lbl.gov} \\
}

\begin{document}
\maketitle

\begin{abstract}
Highway traffic modeling and forecasting approaches are critical for intelligent transportation systems. 
Recently, deep-learning-based traffic forecasting methods have emerged as state of the art for a wide range of traffic forecasting tasks. However, these methods require a large amount of training data, which need to be collected over a significant period of time. 
This can present a number of challenges for the development and deployment of data-driven learning methods for highway networks that suffer from lack of historical data. A promising approach to address this issue is transfer learning, where a model trained on one part of highway network can be adapted for a different part of the  highway network. We focus on diffusion convolutional recurrent neural network (DCRNN), a state-of-the-art graph neural network for highway network forecasting. It models the complex spatial and temporal dynamics of the highway network using a graph-based diffusion convolution operation within a recurrent neural network. DCRNN cannot perform transfer learning, however, because it learns location-specific traffic patterns, which cannot be used for unseen regions of the network. To that end, we develop a new transfer learning approach for DCRNN, where a single model trained on data-rich regions of highway network can be used to forecast traffic on unseen regions of the highway network. We evaluate the ability of our approach to forecast the traffic on the entire California highway network with one year of time series data. We show that \tlDCRNN can learn from several regions of the California highway network and forecast the traffic on the unseen regions of the network with high accuracy. Moreover, we demonstrate that \tlDCRNN can learn from  San Francisco region traffic data and can forecast traffic on the Los Angeles region and vice versa.
\end{abstract}

\section{Introduction}

\input{intro}

\input{method}

\section{Experimental results}

\input{results}

\section{Related work}
\input{rel}

\section{Conclusion and future work}

\input{conc}

\bibliographystyle{unsrt}
\bibliography{references.bib}
\section*{Supplementary document}
\input{supplement}

\end{document}

%% file: intro.tex
With steadily increasing urbanization in major cities in the United States, city planning and state-level department of transportation organizations are now focusing heavily on traffic management systems.  The complexity of traffic management has, in the past, been too high a bar for most control-oriented technology solutions. Signal control, variable messaging signs, and traffic metering onto highways have managed at a localized level in the past and sufficed.  With increased densities of vehicles on the road network, however, these solutions are falling short, and traffic is having a detrimental impact on area economics, productivity, emissions, and quality of life. As an example, the costs associated with congestion in California have been estimated to be on the order of \$29B, with approximately \$1.7B attributed to each of the Los Angeles (LA)  and San Francisco (SFO) regions alone \cite{trip2018}.

New data collection opportunities and advanced computing capabilities are beginning to emerge in transportation systems. Most of these are being deployed in a ``corridor environment,'' where congested sections of major highways are instrumented to collect a variety of traffic measurements. These measurements are fed back to traffic management centers (TMCs) and decisions systems that, with the approval of humans in the loop, provide timely alerts to authorities and control traffic with cost measures that provide the best outcomes for the corridor, such as congestion metrics, fuel efficiency, and emission. 
Beginning in the 1980s, TMCs have spread widely across the United States, with over 280 TMCs operating in the nation today. The decision systems in use by TMCs are a key component of an efficiently operating transportation system and the subject of significant research.

In parallel to the emerging data collection and computing capabilities afforded to TMCs, techniques to better forecast traffic conditions---a foundational component of a TMC's decision system---are also seeing significant improvements to the state of the art. Statistical techniques such as autoregressive  \cite{williams2003modeling} and Kalman filtering \cite{kumar2017traffic} methods are commonly used for time series forecasting. However, these  models work well only for stationary time series data \cite{al2011selecting}. 
Machine learning methods such as artificial neural networks \cite{chan2012neural,karlaftis2011statistical} and support vector machines  \cite{castro2009online, ahn2016highway} have been shown to outperform classical methods on highly nonliner and nonstationary traffic data. Recently, deep learning methods such as deep belief networks~\cite{huang2014deep} and stacked autoencoders~\cite{lv2015traffic}, recurrent neural network variants \cite{ma2015long,fu2016using}, have emerged as promising approaches because of their ability to capture the long-term temporal dependencies. However, they do not model the spatial dependencies of highway network. 
Convolutaional neural networks with recurrent units have been investigated to model the spatial temporal dynamcis of traffic, where the spatial temporal traffic data are converted to a sequence of images and sequence-to-sequence learning is performed on the images \cite{ma2017learning}. However, these methods  pose significant limitations because they violate the fundamental non-Euclidean property of the network data. While the nearby pixels are correlated in images, nearby locations in a highway can be different because they can be on the opposite side of the highway (for example, one going into the city and one coming out of the city).
To that end, Diffusion Convolutional Recurrent Neural Network (DCRNN) \cite{li2017diffusion} has been proposed to overcome the challenges of spatiotemporal modeling of highway traffic networks. It is a  state-of-the-art graph-based neural network that captures spatial correlation by a diffusion process on a graph and temporal dependencies using a sequence-to-sequence recurrent neural network. These data-driven forecasting methods can help  better utilize the transportation resources, develop policies for better traffic control, optimize  routes, and reduce incidents and accidents on the road.



The high performance of the deep learning models including DCRNN can be attributed to the availability of significant amounts of historical data for training. While these deep learning models can be transformational for the performance of a transportation decision system, many U.S. states  do not have the data collection infrastructure that can provide such historical data for training. Moreover, the dynamic nature of transportation systems dictates that even for systems with highly instrumented corridors, other areas of emerging congestion may not have sufficient instrumentation or historical data. Relatedly, while several new data collection infrastructures have been deployed in various states, the time required for data collection can hinder model development and deployment.
In the absence of an installed data collection infrastructure or areas where sensors are not as geospatially dense, probe data collected from GPS and cellphones can be used as a proxy to measure key traffic metrics such as speed. In these cases, model training can become difficult because historical data  either may be unavailable or cannot be accumulated for privacy concerns. These challenges, combined with the improved traffic forecasting capabilities afforded by the DCRNN approach, suggest that identifying methods to deploy models trained in areas rich in historical data to areas with a paucity of data are especially promising. Additionally, if successful, these methods can allow states, cities, and municipalities to more quickly develop improved traffic forecasting capabilities with a significantly smaller infrastructure investment. Expanding the benefits from just TMCs, many other intelligent transportation applications, such as dynamic routing for freight traffic to congestion pricing based on forecast traffic conditions, could also benefit from localized models trained on data sets from more data-rich locations.


Transfer learning is a promising approach to address the data paucity problem. In this approach, a model trained for one task is reused and/or adapted for a related  task. While transfer learning is widely used for image classification, sentiment analysis, and document classification in the text domain \cite{zhuang2019comprehensive, tan2018survey}, it has received less attention in the traffic forecasting domain.
Using transfer learning methods for graph-based highway traffic forecasting  such as DCRNN is not a trivial task. The reason is that graphs have complex neighborhood correlations as opposed to images, which have a relatively simple local correlations as they are samples from the same Euclidean grid domain \cite{zhou2018graph}. Moreover, DCRNN cannot perform transfer learning because it learns the location-specific spatial temporal patterns in the data and requires the same highway network graph for both training and inference. 

To address these issues, we have developed \tlDCRNN, a DCRNN with transfer learning capability. Given a large highway network with historical data, \tlDCRNN partitions the network into a number of regions. At each epoch, region-specific graphs and its corresponding time series data are used to train a single encoder-decoder model using minibatch stochastic gradient descent. Consequently, the location-specific traffic patterns are marginalized, and the model tries to learn the traffic dynamics as a function of graph connectivity and temporal pattern alone. We conduct extensive experiments on the real-world traffic dataset of the entire California road network from the Performance Measurement System (PeMS) administered by the California Department of Transportation (Caltrans). 
PeMS is one of the first system in the United States to instrument and collect data on the highways at scale. The system has been used for operational analysis, planning, and research studies for almost 20 years. Our contributions are as follows:
\begin{itemize}
\item We develop a new graph-partitioning-based transfer learning approach for diffusion convolution recurrent neural network that learns from data-rich regions of the highway network and makes short-term forecasts for unseen regions of the network.   
\item By marginalizing location-specific information with transfer learning, we show that it is feasible to model the traffic dynamics as a function of temporal patterns and network connectivity. 
\item We demonstrate that our proposed transfer learning method can learn from different regions of the California highway network and can forecast traffic on unseen regions. We show the efficacy of the method by learning from the San Francisco (SFO)  region data and forecasting for the Los Angeles (LA) region and vice versa.
\end{itemize}

%% file: method.tex
\section{Problem setup}

The short term highway traffic forecasting problem can be defined on a weighted directed graph $\mathcal{G} = (\mathcal{V}, \mathcal{E}, \mathcal{A})$, where $\mathcal{V}$ is a set of $N$ nodes that represent highway sensor locations, $\mathcal{E}$ is the set of directed edges connecting these nodes, and $\mathcal{A}\in R^{N\times N}$ is the weighted adjacency matrix that represents the connectivity between the nodes in terms of highway network distance. The traffic state at time step $t$ is represented as a graph signal $X_t \in R^{N \times F}$ on the graph $\mathcal{G}$, where $F$ is the number of traffic metrics of interest (e.g., traffic flow, traffic speed, and density that change over time). Given $H$ historical observations of the traffic state $X =(X_{t_1}, X_{t_2}, ..., X_{t_H}) \in \mathds{R}^{H\times N\times F} $ and $P$ observations of the current traffic state $X =(X_{t_1}, X_{t_2}, ..., X_{t_P}) \in \mathds{R}^{P\times N\times F}$ on the graph $\mathcal{G}$, where $H>>P$, the goal is to develop a model that can forecast the traffic state of the next $Q$ time steps on all nodes of the graph, $\hat{Y} = (\hat{X}_{t_{P +1}} , \hat{X}_{t_{P +2}} , ..., \hat{X}_{t_{P +Q}} ) \in \mathds{R}^{Q\times N\times F} $. 

Let $\mathcal{G}' = (\mathcal{V}', \mathcal{E}', \mathcal{A}')$ be the graph with $N'$ nodes that represents the highway network for which we do not have the historical time series data. 
Given $P$ observations of the current traffic state $X' =(X'_{t_1}, X'_{t_2}, ..., X'_{t_P}) \in \mathds{R}^{P\times N'\times F}$ on the graph $\mathcal{G'}$, the goal is to develop a model that can forecast the traffic state of the next $Q$ time steps on all nodes of the graph $\mathcal{G}'$, $\hat{Y}' = (\hat{X}'_{t_{P +1}} , \hat{X}'_{t_{P +2}} , ..., \hat{X}'_{t_{P +Q}} ) \in \mathds{R}^{Q\times N'\times F}$.

In the context of Pan and Yang's transfer learning classification \cite{pan2009survey}, our problem setup corresponds to the transductive transfer learning setting, where the source and target tasks are the same (short-term traffic forecasting on graphs), while the source and target (unseen regions of the highway network) domains are different but related.

\section{Diffusion Convolution Recurrent Neural Network (DCRNN)}\label{dcrnn}

DCRNN is a state-of-the-art method for short term traffic forecasting \cite{li2017diffusion}. It is an encoder-decoder neural network architecture that performs sequence-to-sequence learning to carry out multistep traffic state forecasting. A simple and powerful variant of recurrent neural networks, called gated recurrent units (GRUs) \cite{cho2014learning} is used to design the encoder-decoder architecture. The matrix multiplications in GRUs is replaced with a diffusion convolution operation to make the DCRNN cell. In an $L$ layered DCRNN architecture, each layer  consists of $R$ number of DCRNN cells.
The DCRNN cell is defined by the following set of equations:
\begin{equation*}
\begin{array}{lcl}
r^t & = &  \sigma (W_{r\bigstar \mathcal{G}} [X_{t}, h_{t-1}] + b_r) \\
u^t & = &  \sigma (W_{u\bigstar \mathcal{G}} [X_{t}, h_{t-1}] + b_u) \\
c^t & = &  tanh (W_{c\bigstar \mathcal{G}} [X_{t} (r_t \odot h_{t-1})] + b_c) \\
h_t & = & u_t \odot h_{t-1} + (1 - u_t) \odot c_t ,
\end{array}
\end{equation*}
where $X_{t}$ and $h_t$ denote the input and final state
at time $t$, respectively; $r_t$, $u_t$, and $c_t$ are the reset gate, update gate, and cell state at time $t$, respectively; $\bigstar G$ denotes the diffusion convolution; and $W_r, W_u, and W_c$  are parameters for the corresponding filters. The diffusion convolution $\bigstar G$
operation over the input graph signal $X$ and convolution filter $W$, which learns the representations for graph-structured data during training, is defined as
\begin{equation}\label{eq_1}
 W_{\bigstar \mathcal{G}} X = \sum_{d=0}^{K-1} (W_O (D_O^{-1}\mathcal{A})^d + W_{I}(D_I^{-1}\mathcal{A})^d) X ,
\end{equation}
where $K$ is a maximum number of diffusion steps; $D_O^{-1}\mathcal{A}$ and $D_I^{-1}\mathcal{A}$ are transition matrix of the diffusion process and the reverse one, respectively; $D_O$ and $D_I$ are the in-degree and out-degree diagonal matrices, respectively; and $W_O$ and $W_I$ are the learnable filters for the bidirectional diffusion process. The in-degree and out-degree diagonal matrices provide the capability to capture the effect of the upstream as well as the downstream traffic. The driving distances between sensor locations are used to build the adjacency matrix $\mathcal{A}$; a Gaussian kernel \cite{shuman2012emerging} and a threshold $\tau$ parameter are used to sparsify $\mathcal{A}$.

During the training of DCRNN, a minibatch of time series sequence,  each of length $P$ from historical time series data $X$, is given as an input to the encoder. The decoder receives a fixed-length hidden representation of the data from the encoder and forecasts the next $Q$ time steps for each sequence in the minibatch. The layers of DCRNN are trained by using backpropagation through time. DCRNN learns the weight matrices in Equation \ref{eq_1} by minimizing the mean absolute error (MAE) as a loss function.

\section{Transfer learning DCRNN }
\label{sec_tlDCRNN}
Similar to the convolution operation on images, the diffusion convolution $\bigstar G$ on a graph is designed to capture patterns that are local to a given node \cite{atwood2016diffusion}. This operation learns the latent representation of a diffusion process that starts from a given node to the neighboring connected node, and it is particularly suitable to capture the local diffusion behavior of traffic dynamics. From Equation \ref{eq_1}, we see that DCRNN becomes location-specific because of the presence of the weighted adjacency matrix $\mathcal{A}$ in the diffusion step. As a result, the convolution filter $W$ is dependent on the given $\mathcal{A}$, which is kept constant throughout the training process. Our hypothesis is that if we change the graph $\mathcal{A}$ and the corresponding time series data during the training process, then we can make diffusion convolution filters generic as opposed to location-specific. 
Consequently, the resulting model is more generalizable and can be used to forecast traffic on unseen graphs.

\begin{figure}
  \centering
    \includegraphics[width=0.9\textwidth]{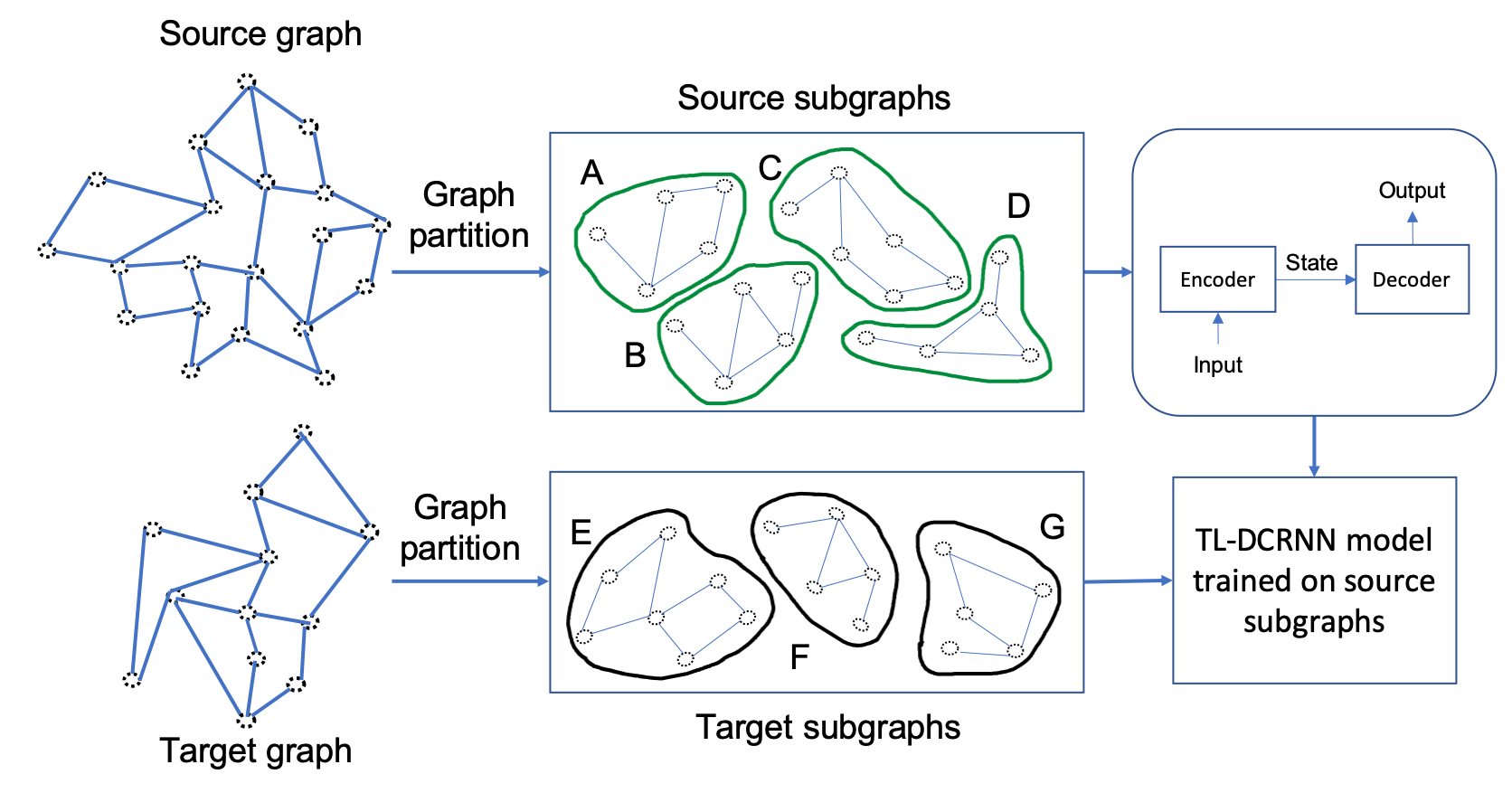}
  \caption{\tlDCRNN partitions the large highway network with historical data into a number of subgraphs using a graph partitioning method. These subgraphs are used to train the the encoder-decoder architecture with diffusion convolution recurrent neural network cells. Given an unseen graph during inference, \tlDCRNN partitions the graph and uses the trained model for short-term traffic forecasting.}
  \label{fig_method_transfer}
\end{figure}

A high-level overview of the proposed \tlDCRNN is shown in Figure \ref{fig_method_transfer}.
Given the source graph $\mathcal{G}$ with the historical data $X$, \tlDCRNN first partitions it into a $m$ subgraphs with equal numbers of $n$ nodes using a graph partitioning method that takes only the weighted adjacency matrix $\mathcal{A}$ of $\mathcal{G}$. Let $\mathcal{G}_p = \{\mathcal{G}_{p}^{1}, \ldots, \mathcal{G}_{p}^{m}\} = \{(\mathcal{V}_p^1, \mathcal{E}_p^1), \ldots, (\mathcal{V}_p^{m}, \mathcal{E}_p^{m})\}$ and $X_p = \{X_{p}^{1}, \ldots, X_{p}^{m}\}$ are the set of $p$ subgraphs and their corresponding time series data. The minibatch stochastic gradient update of \tlDCRNN is same as that of DCRNN, where for a given $\mathcal{G}_{p}^{i}$, a batch of input and output time series sequence from $X_{p}^{i}$ is used to compute the errors and update the weights of the encoder and decoder architecture by backpropagation through time. The minibatch is constructed to preserve time ordering, a common approach in sequence-to-sequence time series modeling. The subgraph epoch for a given $\mathcal{G}_{p}^{i}$ uses all the data in $X_{p}^{i}$ as a series of minibatches to update the parameters of the encoder decoder architecture. The epoch  runa a subgraph epcoh for each $\mathcal{G}_{p}^{i} \in \mathcal{G}_p$.


For inference, \tlDCRNN partitions the target graph $\mathcal{G'}$ into $m'$ subgraphs $\{\mathcal{G'}_{p}^{1}, \ldots, \mathcal{G'}_{p}^{m'}\}$ such that each subgraph $\mathcal{G'}_{p}^{j}$ has $n$ nodes using the graph partitioning method. Given the current state of the traffic as a sequence of $P$ time series on a subgraph $\mathcal{G'}_{p}^{j}$, the \tlDCRNN trained model forecasts the traffic for the next $Q$ time steps.

From the learning perspective, the key difference between DCRNN and \tlDCRNN is that the former learns the location-specific spatial temporal patterns on a static graph whereas the latter marginalizes the location-specific information and learns the spatial temporal patterns across multiple subgraphs. Consequently, the model weights are trained in such way that it can provide generalization to similar but unseen graphs, a capability that can be used for forecasting on a unseen highway network graph.

\subsubsection*{Note on graph partitioning:}
Several graph partitioning methods exist that can be leveraged for partitioning the graphs in the training and inference phase. Typically, these methods cannot partition the graph into equal-sized subgraphs, but they can give partitions of similar sizes. In these cases, we can find the largest size of the graph and perform zero padding for all other subgraphs. The unseen graph $\mathcal{G'}$ does not need to be as large as $\mathcal{G}$. For a given $\mathcal{G'}$, when $N'$ modulo $n$ value is relatively smaller than $n$, one can use zero padding to the subgraph whose size is smaller than $n$. When the modulo value is much smaller than $n$, the method of subgraphs with overlapping nodes \cite{mallick2019graphpartitioningbased} can be adopted, where a subgraph includes nodes from neighboring geographically close subgraphs.

\subsection*{Assumptions and implications}
We assume that the graph $\mathcal{G}$ represents a large highway network such that it is amenable to graph partition and, in particular, the partitioned subgraphs expose a range of traffic dynamics. Otherwise, the transfer learning ability of the method will suffer. Furthermore, the transductive transfer learning assumptions of related domain apply to our method. The highway traffic data can be seen as the samples generated from a high-dimensional distribution parameterized by highway vehicle composition, infrastructure, and entry exit dynamics, among others. Given completely different distribution sample, the model will not be able to perform transfer learning with high accuracy. For example, a model that is trained on certain regions of California  will not generalize to highway network traffic forecasting in completely different regions of the world such as China and India, where the highway vehicle composition, infrastructure, and  traffic dynamics can be dramatically different.

%% file: results.tex
\label{sec_exp}


Our dataset has 11,160 sensors for the entire year of 2018.  Besides the time series data, PeMS  captures spatial information such as the latitude and longitude of each station. 
We computed the pairwise driving distances between the sensors using the latitude and longitude and built the adjacency matrix using a thresholded Gaussian kernel as prescribed in \cite{li2017diffusion}. 

To generate subgraphs for the \tlDCRNN training, we partitioned the whole California highway traffic network graph into 64 roughly equal-sized subgraphs using Metis's $k$-way partitioning method \cite{metis}. Previously, we had shown that instead of training the full California network, decomposing the large graph into 64 subgraphs and training 64 DCRNNs independently results in better accuracy \cite{mallick2019graphpartitioningbased}. We divided the 64 subgraphs into two equal sets: source subgraphs  (numbered from 1 to 32) and target subgraphs (numbered from 33 to 64). The 32 subgraphs in the source subgraphs  were selected systematically to ensure uniform coverage over entire California network: the first 8 subgraphs were selected from eight  districts or locations of California:  North Central, Bay Area, South Central, Los Angeles, San Bernardino, Central, San Diego, and Orange County. We randomly selected 16 subgraphs from the remaining subgraphs without replacement. All the unselected subgraphs were grouped in the target subgraphs. 
From   one  year of  the time series data,  we  use   70\% of  the  data  ($\approx$36  weeks)  for  training, 10\%  ($\approx$5 weeks) for  validation, and  20\%  ($\approx$10  weeks)  for  testing.
Consequently, we have six datasets. See Table \ref{tab_dataset_definition} for the dataset summary and nomenclature adopted. 

\begin{table}
    \caption{Datasets used for the experiments: 64 subgraphs are divided into source and target subgraphs. The source set ({\ttfamily src}) contains subgraphs from 1 to 32 partitions, and the target set ({\ttfamily tgt}) contains subgraphs from 33 to 64 partitions. The timelines of training, validation, and testing are given in the description.}
    \label{tab_dataset_definition}
    \centering
        \begin{tabular}{l|p{9.3cm}}
    \multicolumn{1}{c}{\bf{Dataset}} & \multicolumn{1}{c}{\bf{Description}} \\ \hline
    \multicolumn{2}{c}{source subgraphs ($1 \ldots 32$)}\\ \hline
        \trainss & $\approx$36  weeks of time series data (1 Jan. 2018 to 13 Sept. 2018) \\
        \valss & $\approx$5  weeks of time series data (13 Sept. 2018 to 20 Oct. 2018) \\
        \testss & $\approx$10  weeks of time series data (20 Oct. 2018 to 31 Dec. 2018) \\ \hline
        \multicolumn{2}{c}{target subgraphs ($33 \ldots 64$)}\\ \hline
        \traints & $\approx$36 weeks of time series data (1 Jan. 2018 to 13 Sept. 2018) \\
        \valts & $\approx$5  weeks of time series data (13 Sept. 2018 to 20 Oct. 2018) \\
        \testts & $\approx$10  weeks of time series data (20 Oct. 2018 to 31 Dec. 2018) \\ \hline
        \end{tabular}
\end{table}

Because of the fixed dimension of the input and adjacency matrix used in the \tlDCRNN, all subgraphs should contain an equal number of nodes.  We added rows and columns filled with zeros to make all the adjacency matrices exactly equal in size. 

We refer the reader to the supplement page of the paper for hardware configurations, software environments, library versions, hyperparameters settings, data preparation, and data prepreprocessing methods. We note that the hyperparameter values were set as default values for the open-source DCRNN implementation \cite{li2017diffusiongit}.

We used speed as the traffic forecasting metric. For training and inference, the forecast horizons were set to 60 minutes, respectively: the encoder gets 60 minutes of traffic data (time and speed) (12 observations, one for every five minutes), and the decoder outputs the forecasts for the next 60 minutes (12 predictions, one for every five minutes). MAE was used as the training loss and test accuracy metric for comparing the  methods.


We computed pairwise MAE difference values because the MAE values of $M1$ and $M2$ are computed on the same set of nodes. We concluded that $M2$ is better than $M1$. We adopted one-sided null hypothesis that the median difference between the two MAE distributions is greater than or equal to zero (MAE value from $M2$ is lower than or similar to that of $M1$). We computed $p-$value from the test and rejected  the null hypothesis when the $p$-value was less than $0.05$ (at a confidence level of 5\%) in favor of the alternative that the median is less than zero (MAE value from $M1$ is lower than those of $M2$). 

\subsection{Impact of number of subgraphs and epochs} \label{sec_impact_parti}
We also conducted an exploratory analysis to study the impact of the number of subgraphs and epochs on the forecast accuracy. We show in this section that \tlDCRNN benefits from the adoption of a large number of subgraphs and epochs for training.

We used three values for the number of subgraphs---8 (1 to 8), 16 (1 to 16), and 32 (1 to 32)---and their corresponding time series data. All models were trained for 1 epoch on \trainss. Approximately 6 minutes were needed to train each  subgraph in each epoch. Therefore, for 8, 16, and 32 subgraphs, training took approximately 48, 96, and 192 minutes, respectively.  

\begin{figure}
  \centering
    \includegraphics[width=0.9\textwidth]{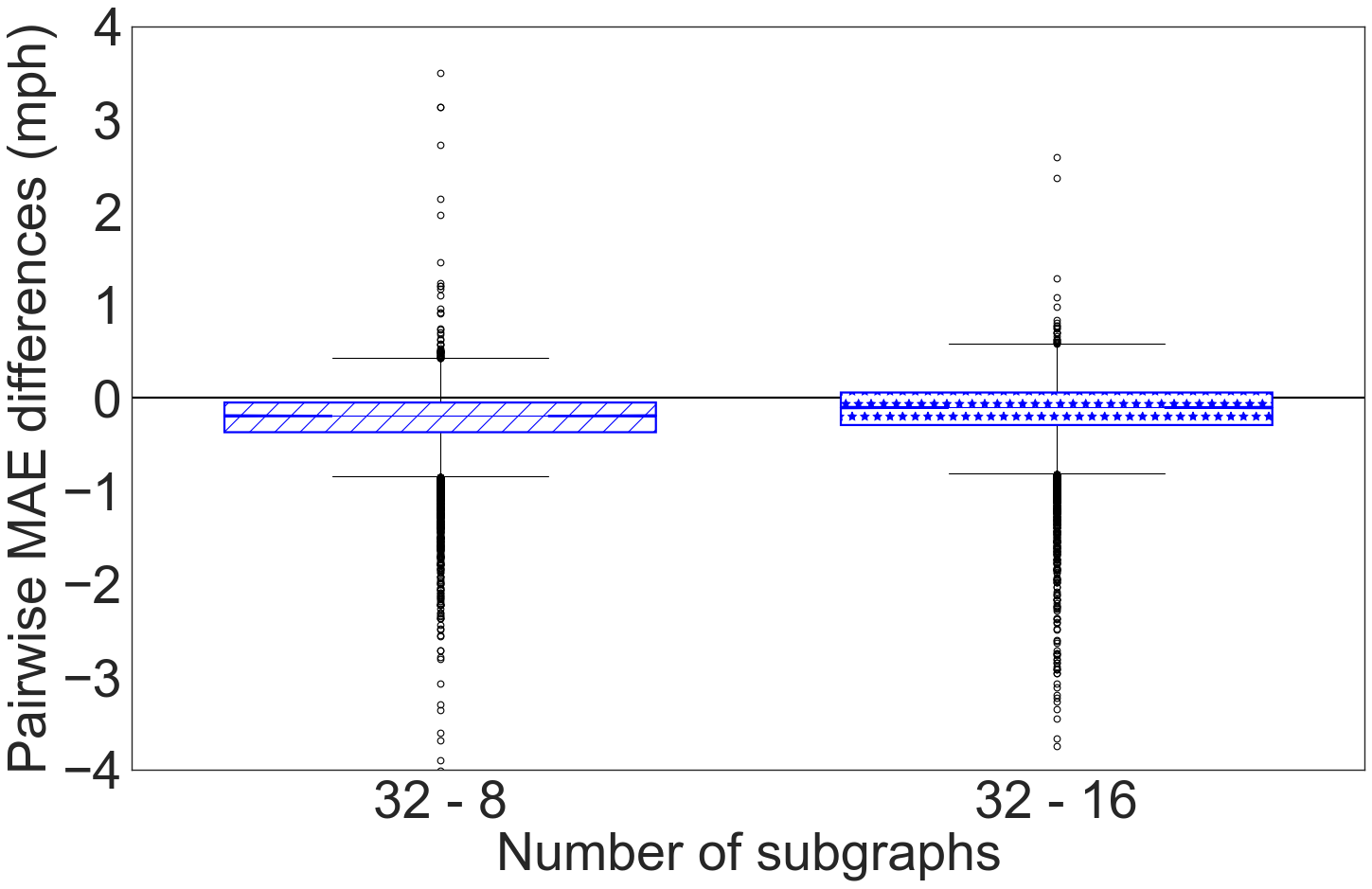}
  \caption{Distributions of pairwise MAE differences between \tlDCRNN models trained with 32 and 8 subgraphs ($32 - 8$) and ($32 - 16$)  subgraphs. Values below zero indicate that the MAE value of the 32 subgraphs is lower than that of the 8 and 16 subgraphs.}
  \label{fig_1e}
\end{figure}

Using box-whisker plots, we showed in Figure \ref{fig_1e} the pairwise MAE difference between 32 and 8 and 32 and 16 subgraphs on 5,579 nodes in \valss from the three models . A difference value below zero indicates that 32 subgraph MAE is lower than the other.   
We can see that 32 subgraph MAE values are lower than those of 8 and 16. We also compared the MAE values obtained from 8 and 16 subgraphs with the one-sided paired Wilcoxon test,. The results show that the latter is better than the former ($p-$value of $1.28 \times 10^{-252}$). We found that 4,051 out of 5,579 nodes have lower MAE values when increasing the subgraphs from 8 to 16. 
Similarly, the one-sided Wilcoxon signed-rank test confirmed the improvement in MAE values when 32 subgraphs are adopted for training ($p-$value of $4.80 \times 10^{-282}$). 
We observed that 3,686 out of 5,579 nodes in the \valss have lower MAE values when increasing the number of subgraphs from 16 to 32. Therefore, we adopted all 32 subgraphs for training (5,579 nodes) for the rest of the experiments.

\begin{figure}
  \centering
    \includegraphics[width=0.9\textwidth]{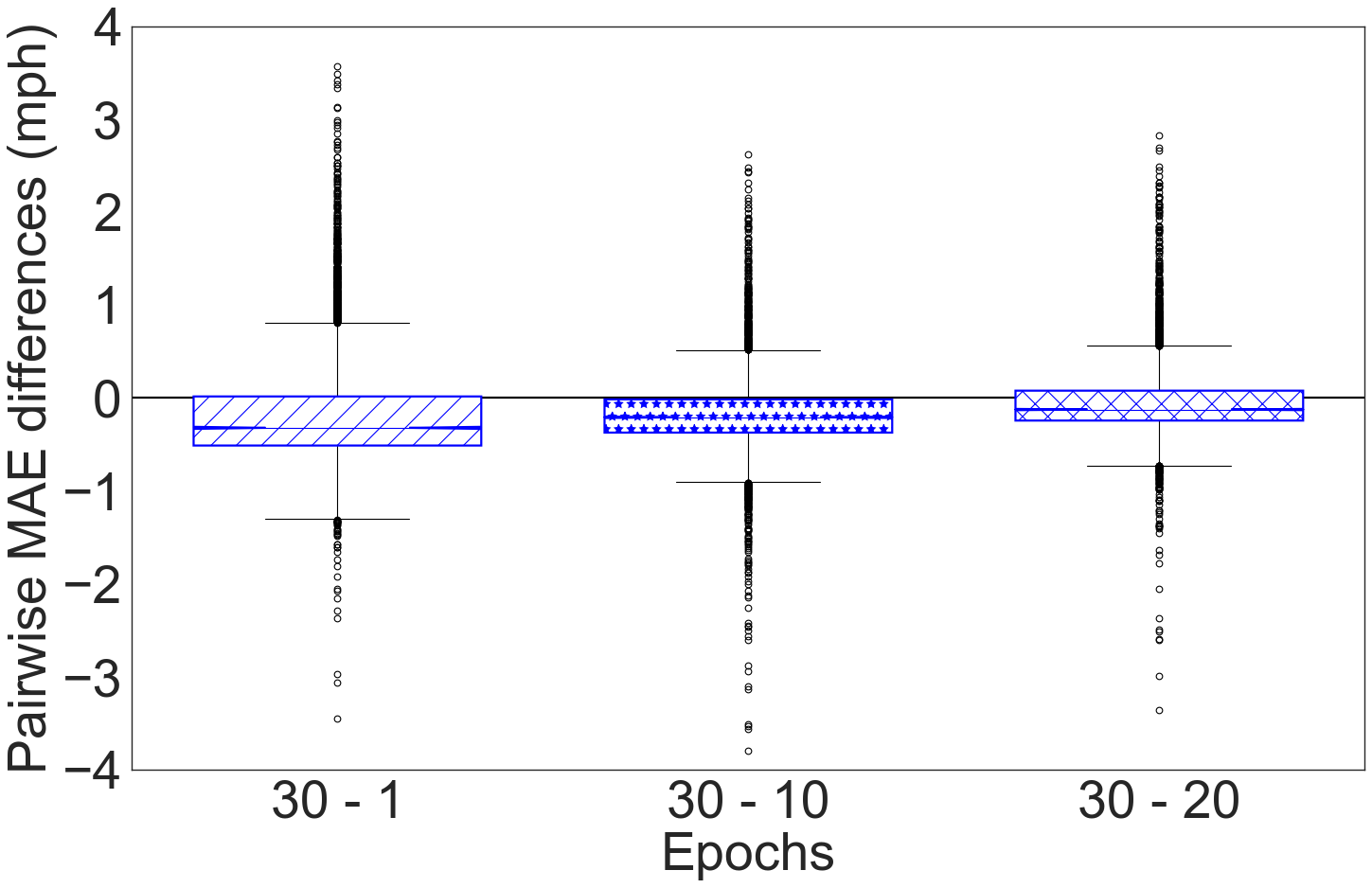}
  \caption{Distributions of pairwise MAE difference between \tlDCRNN models trained with 30 and 1 epochs ($30-1$), 30 and 10 ($30-10$), and 30 and 20 epochs ($30-20$). Values below zero indicate that MAE values of 30 epochs are lower than the others.} 
  \label{fig_10e}
\end{figure}

Next, we increased the number of training epochs of \tlDCRNN and analyzed its impact on forecasting accuracy. We trained \tlDCRNN on \trainss for 1, 10, 20, and 30 epochs; the model training took approximately 192, 1920, 3840, and 5760 minutes, respectively. 
We evaluated the trained model on \valss. The distributions of the pairwise MAE difference between 30 and 1, 30 and 10, and 30 and 20 epochs for 5,579 nodes are shown in  Figure \ref{fig_10e}. 
The paired one-sided Wilconxon test obtained $p-$values of  $4.95 \times 10^{-33}$,  $4.08 \times 10^{-283}$, and $2.14 \times 10^{-18}$ for the comparison of 1 and 10, 10 and 20, and 20 and 30 epochs, respectively, confirming that the null hypothesis can be rejected at a confidence level of 5\%. For the rest of the experiments,we therefore adopted 30 epochs for training.

\subsection{Transfer learning on California network}\label{sec_tl_cali}
Here, we showed that the proposed subgraph training strategy of \tlDCRNN for transfer learning is effective.

We trained and tested \tlDCRNN on \trainss and \testts data, respectively. We compared this approach with DCRNN, where we trained a subgraph-specific model for each subgraph in \trainss. This resulted in 32 subgraph-specific trained models. We refer to this method as \ssDCRNN. %
We adopted three test modalities for \ssDCRNN. First, we evaluated each of the 32 \ssDCRNN models on the \valts and selected the one with the lowest MAE for testing on \testts (single model for testing referred as  \ssDCRNN-S). Second, for each subgraph in \testts, we evaluated the 32 \ssDCRNN models on the \valts and selected the model with the lowest MAE for testing on \testts (multiple models for testing referred as \ssDCRNN-M).  Third, we adopted a bagging approach, where  forecasting for each subgraph in \testts is given by the average of 32 model forecasts (bagged models for testing are referred to as \ssDCRNN-B). Note that \ssDCRNN-S and \ssDCRNN-M introduce a bias in favor of \ssDCRNN since the model selection is made on the data from the target subgraphs

\begin{figure}
  \centering
    \includegraphics[width=0.9\textwidth]{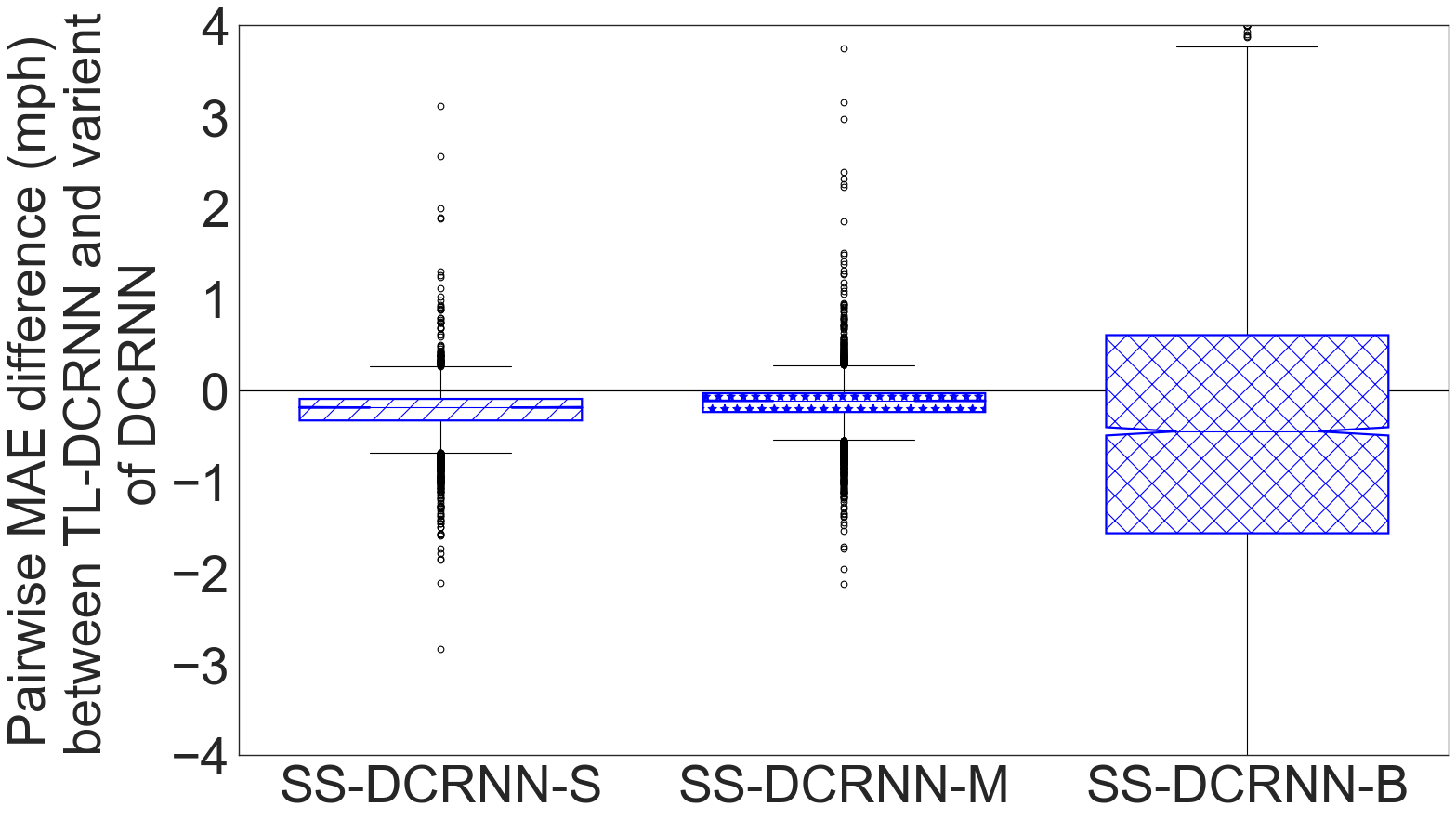}
  \caption{Distributions of pairwise MAE differences between \tlDCRNN  and \ssDCRNN variants for transfer learning on \testts. Values below zero indicate that MAE values obtained by \tlDCRNN are lower than those of \ssDCRNN variants.}
  \label{fig_whole}
\vspace{-0.5cm}
\end{figure}

Figure \ref{fig_whole} shows the distributions of the pairwise MAE differences obtained on \testts by different methods. We  observe that \tlDCRNN outperforms the \ssDCRNN variants. The results show that the MAE values obtained by \tlDCRNN are lower than \ssDCRNN-S, \ssDCRNN-M, and \ssDCRNN-B, on 5,188, 4,577, and 3,416 out of 5,581 nodes, respectively. 
The observed differences are significant according to a paired one-sided Wilcoxon signed-rank test, which showed $p-$values of $0.0$ (\tlDCRNN vs \ssDCRNN-S), $0.0$ (\tlDCRNN vs \ssDCRNN-M), and $2.54 \times 10^{-88}$ (\tlDCRNN vs \ssDCRNN-B).

Given that \tlDCRNN and \ssDCRNN differs only with respect to the model training, the superior performance can be attributed to the proposed subgraph-based training strategy. 
The comparison between \ssDCRNN-B  and \ssDCRNN-M shows that the MAE values obtained by \ssDCRNN-M is lower than those of \ssDCRNN-S on 3,095 out of 5,581 nodes. This is because the subgraph-specific model selection results in a collection of models, which provide more robust forecasts than a single model. 
The MAE values obtained by \ssDCRNN-B are worse than those  of \tlDCRNN, \ssDCRNN-B, and \ssDCRNN-M. Unlike \ssDCRNN-S and \ssDCRNN-M, subgraph-specific models in \ssDCRNN-B do not have any advantage on unseen graphs. While bagging good models increases the accuracy, the same approach with poor models can significantly degrade the accuracy \cite{opitz1999popular}.

\subsection{Direct learning on California network} 
\label{sec_comp}
Here, we showed that when the same subgraphs are used for training and testing, \ssDCRNN achieves results that are better than those of \tlDCRNN.

We trained \ssDCRNN and \tlDCRNN on \trainss and tested them on \testss. The \tlDCRNN model does not learn subgraph-specific traffic dynamics, resulting in one model, whereas \ssDCRNN learns subgraph-specific traffic dynamics, resulting in 32 models.

\begin{figure}
  \centering
    \includegraphics[width=0.9\textwidth]{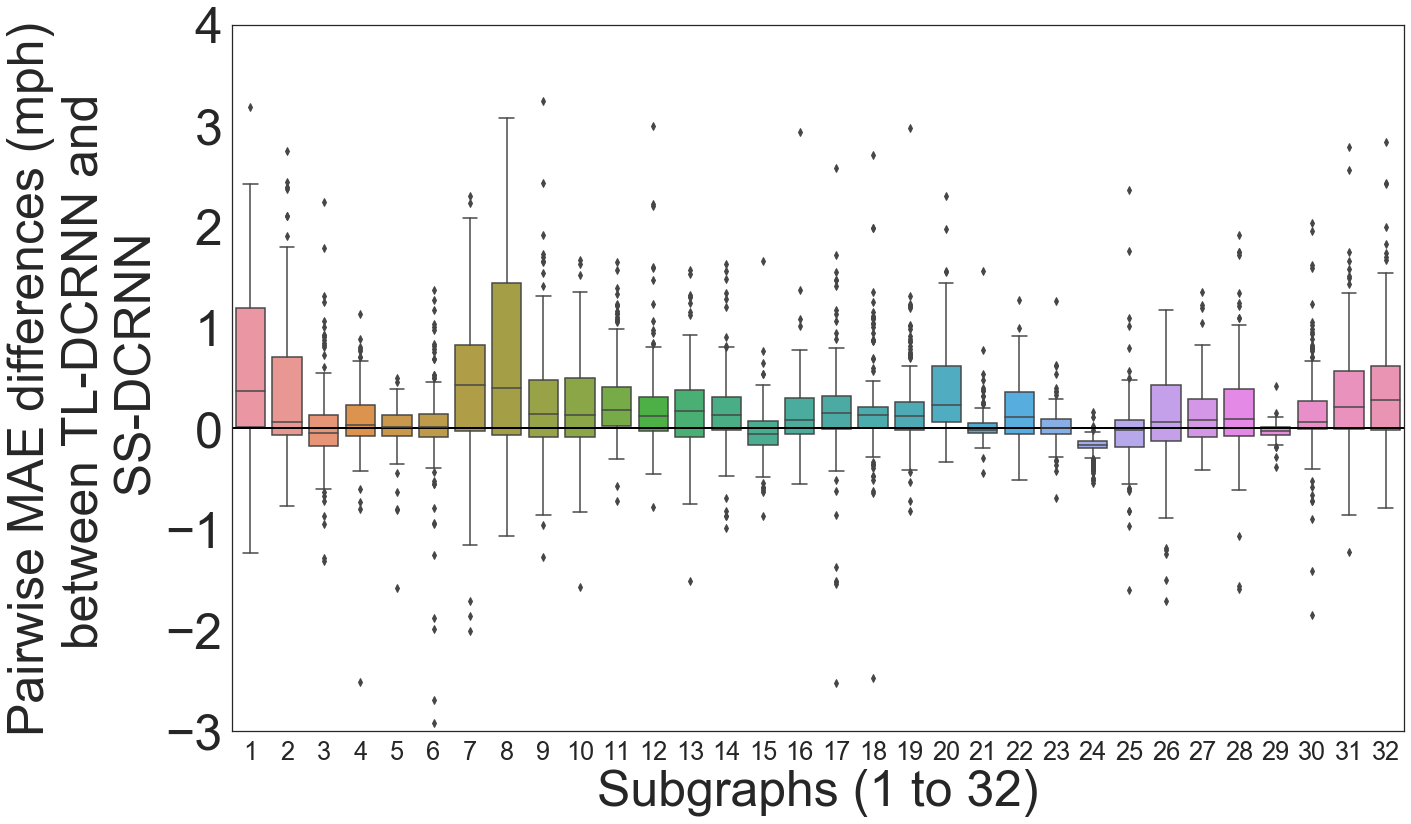}
  \caption{
  Distributions of pairwise MAE differences between \tlDCRNN  and \ssDCRNN for direct learning on \testss. The results are shown for all 32 partitions in \testss. Values greater than zero indicate that MAE values of \ssDCRNN are lower than those  of \tlDCRNN.}
  \label{fig_direct_learning}
\end{figure}

Figure \ref{fig_direct_learning} shows the distribution of pairwise  differences between the MAE values of \tlDCRNN and \ssDCRNN obtained on the 32 partitions in \testss. 
Out of 32 subgraphs, the  medians of the distributors are positive for 26 subgraphs. The results show that the MAE values obtained by \ssDCRNN are lower than those obtained by \tlDCRNN, on 3,412 out of 5,579 nodes in the \testss data. The observed differences are significant according to a paired one-sided Wilcoxon signed-rank test, which showed a $p-$value of $1.72 \times 10^{-142}$. 
Superior performance of the 32 models of \ssDCRNN can be attributed to the fact that they take into account the location-specific information for learning the traffic dynamics. In single-model \tlDCRNN the location-specific spatiotemporal learning is traded off to achieve transfer learning capability. In particular, the inductive bias due to location-specific training results in generalization, which is stronger than that of the transfer learning \ssDCRNN under the adopted experimental settings.

\subsection{Transfer learning between LA and SFO}
In Section \ref{sec_tl_cali} we selected the subgraphs for training to have uniform coverage on the districts of California state. Here, we compare \tlDCRNN and \ssDCRNN on the two major districts of California and demonstrate that \tlDCRNN model trained on LA (SFO) can be used for forecasting SFO (LA). 

In the PeMS system, the LA and SFO districts have 2,716 and 2,382 sensor locations, respectively.  We partitioned the highway traffic network of LA and SFO into 15 and 13 subgraphs, respectively, to keep the number of nodes per partition as close as 64 partitions used in Section \ref{sec_tl_cali}. We trained two \tlDCRNN models on \trainss of LA and SFO and tested  them on \testts of SFO and LA, respectively. We used the same timeline for training, validation, and testing as shown in Table \ref{tab_dataset_definition}. %
We adopted the three test modalities for \ssDCRNN (as discussed in Section \ref{sec_tl_cali}): \ssDCRNN-S, \ssDCRNN-M, and \ssDCRNN-B. Moreover, we included  direct  learning in the analysis, where \ssDCRNN was trained and tested on the same subgraphs of LA (SFO).

\begin{figure}
  \centering
    \includegraphics[width=0.9\textwidth]{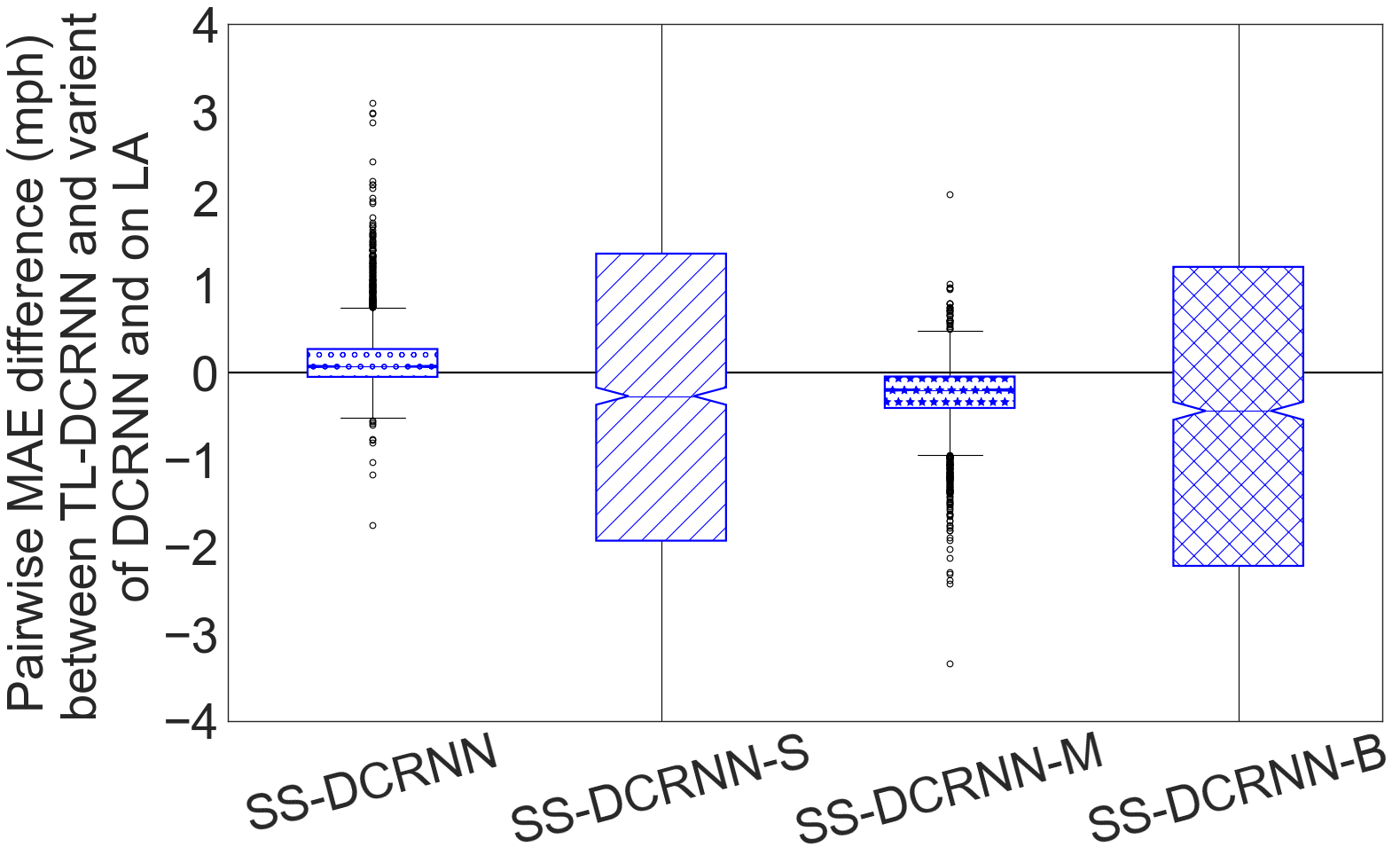}
  \caption{Distributions  of  pairwise  MAE  differences  between \tlDCRNN and \ssDCRNN variants  for  transfer  learning  on the LA region.  Values  below  zero  indicate  that  MAE  values  obtained by \tlDCRNN are lower than those of  all variants of \ssDCRNN except direct learning using \ssDCRNN.}
  \label{fig_LA}
\end{figure}

\begin{figure}
  \centering
    \includegraphics[width=0.9\textwidth]{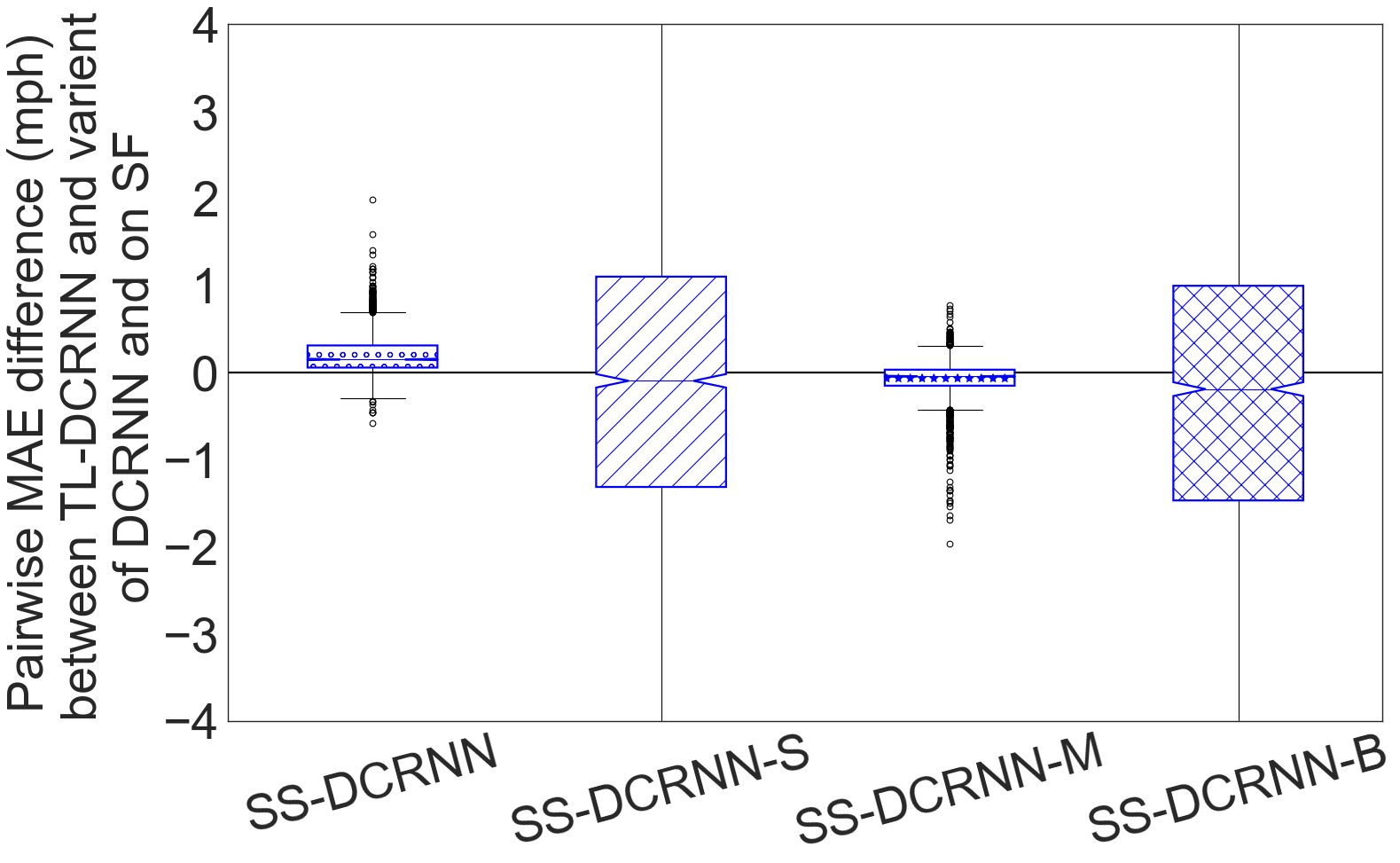}
  \caption{Distributions  of  pairwise  MAE  differences  between \tlDCRNN and \ssDCRNN variants  for  transfer  learning  on SFO region.  Values  below  zero  indicate  that  MAE  values  obtained by \tlDCRNN are lower than those obtained for all variants of \ssDCRNN except direct learning using \ssDCRNN.}
  \label{fig_SF}
\end{figure}

Figure \ref{fig_LA} shows the distributions of the pairwise MAE differences between \ssDCRNN methods and \tlDCRNN when the models were trained on SFO and tested on LA (except direct learning DCRRN, which was trained on LA).
We  observe that \tlDCRNN outperforms \ssDCRNN-S, \ssDCRNN-M, and \ssDCRNN-B. For LA, the MAE values obtained by \tlDCRNN are lower than those obtained by \ssDCRNN-S, \ssDCRNN-M, and \ssDCRNN-B on 1,467, 2,195, and 1,553 out of 2,717 nodes, respectively. The paired one-sided Wilcoxon signed-rank test shows $p-$values of $4.96 \times 10^{-09}$ (\tlDCRNN vs \ssDCRNN-S), $7.40 \times 10^{-260}$ (\tlDCRNN vs \ssDCRNN-M), and $1.85 \times 10^{-22}$ (\tlDCRNN vs \ssDCRNN-B). The trend is similar for the results on SFO,  shown in Figure \ref{fig_SF}. The MAE values obtained by \tlDCRNN are lower than those of \ssDCRNN-S, \ssDCRNN-M, and \ssDCRNN-B, on 1,237, 1,546, and 1,297 out of 2,383  nodes, respectively. The observed differences are significant according to a paired one-sided Wilcoxon signed-rank test, which shows $p-$values of $0.0007$ (\tlDCRNN vs \ssDCRNN-S), $2.62 \times 10^{-68}$ (\tlDCRNN vs \ssDCRNN-M), and $8.85 \times 10^{-10}$ (\tlDCRNN vs \ssDCRNN-B). 

Similar to the results in Section \ref{sec_comp}, direct learning \ssDCRNN obtains MAE values lower than those of \tlDCRNN. 

\begin{figure}
  \centering
    \includegraphics[width=0.9\textwidth]{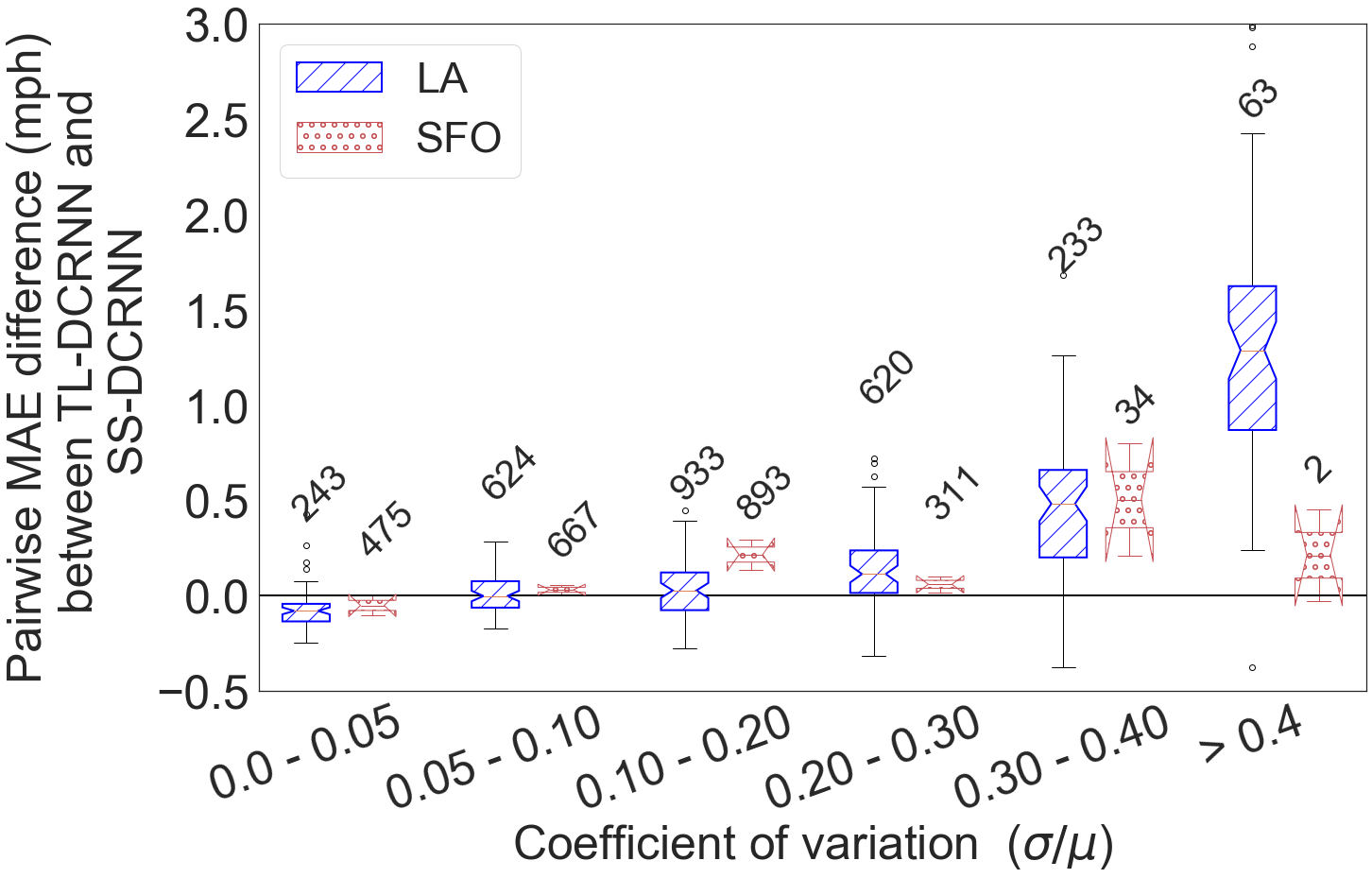}
  \caption{Pairwise MAE differences between \tlDCRNN and \ssDCRNN and with respect to coefficient of variation values computed on \testts on LA and SFO regions. The numbers above each box show the number of observations (nodes) in the distribution.}
  \label{fig_LA1}
\end{figure}

To gain further insight into the observed \tlDCRNN errors, we computed, for each node, the coefficient of variation given by the ratio of standard deviation and mean of the time series in \testts. This measure can be used as a proxy to measure the traffic dynamics: smaller values indicate that the speed is stable (less dynamic), and larger values mean that a wide range of speed values have been observed (more dynamic). Figure \ref{fig_LA1} shows the distribution of pairwise MAE differences between \tlDCRNN vs \ssDCRNN as a function of coefficient of variation intervals. On LA data, we can see a clear trend in which an increase in the coefficient of variation values increases the MAE difference values. \ssDCRNN forecasts become more accurate than that of \tlDCRNN. This is becasue \ssDCRNN with 32 models trained and tested on the same graph captures the location-specific traffic dynamics better than  does \ssDCRNN with single model that was trained on the SFO dataset. We can observe a similar trend in the SFO dataset as well. The significant drop in the pairwise MAE differences for coefficient of variation larger than 0.4 can be attributed to small number (2) of nodes. The number of nodes with coefficient of variation larger than 0.3 is rather small ($36$) when compared to the LA dataset. This shows that the SFO traffic is less dynamic when compared with LA traffic. 

\subsection{Comparison with other methods}
In this section, we present our comparison of \tlDCRNN with other short-term traffic forecasting methods proposed in the literature. We show that despite not being trained on the same highway network graph and data, \tlDCRNN achieves an accuracy that is better than or comparable to the accuracy of the other traffic forecasting methods.

Specifically, we compared \tlDCRNN with the following methods: (1) autoregressive integrated moving average
(ARIMA) \cite{makridakis1997arma}, which considers only the temporal relationship of the data; (2) support vector
regression (SVR) \cite{wu2004travel}, a linear support vector machine for forecasting, (3) a feed-forward
neural network (FNN) \cite{raeesi2014traffic} with two hidden layers, (4) fully connected LSTM (FC-LSTM) \cite{sutskever2014sequence} with encoder-decoder architecture, (5) a spatiotemporal graph convolutional network (STGCN) \cite{yu2017spatio}, which combines graph convolutions and gated temporal convolutions, (6) a diffusion convolutional recurrent neural network (DCRNN),  as discussed in Section \ref{dcrnn}, (7) Graph waveNet \cite{wu2019graph}, a CNN-based method with stacked dilated casual convolutions for handling temporal dependencies, and (8) a graph multi-attention network (GMAN) \cite{zheng2019gman}, an  encoder-decoder architecture with multiple spatio-temporal attention. For this comparison, we used the accuracy results reported in \cite{zheng2019gman}, where GMAN results are compared with other methods.

All the methods were benchmarked on the PEMS-BAY dataset with 325 sensors in the Bay Area with 6 months of time series data ranging from Jan. 1, 2017 to June 30, 2017. We used 70\% of the data for training (Jan. 1, 2017, to May 7, 2017), 10\% of the data was used for validation (from May 7, 2017, to May 25, 2017), and 20\% was used for testing (from May 25, 2017, to June 30, 2017).
We trained \tlDCRNN on the LA dataset on the same timeline and tested on the PEMS-BAY dataset.

In addition to MAE, we used root mean wquare error (RMSE) and mean absolute percentage error (MAPE) metrics to compare the accuracy of these models. The comparison results are shown in Table \ref{tab_comp}. We observe that \tlDCRNN achieves MAE of 2.13, which is better than ARIMA (3.38), SVR (3.28), FNN (4.46), FC-LSTM (2.37), and STGCN (2.49). The trend is similar for RMSE and MAPE. Although  DCRNN, Graph Wevenet, and GMAN were trained on the PEMS-Bay dataset, their accuracy metrics were not significantly better than those obtained by \tlDCRNN trained on the LA dataset.


\begin{table}
\centering
\caption{Accuracy metrics comparison of TL-DCRNN with other short-term traffic forecasting approaches.}
\begin{tabular}{|l|l|l|l|}
\hline
\bf{Method} & \bf{MAE} & \bf{RMSE} & \bf{MAPE} \\ \hline
\multicolumn{4}{|c|}{Training and testing on PEMS-BAY}\\ \hline
ARIMA & 3.38 & 6.50 & 8.30\% \\ \hline
SVR & 3.28 & 7.08 & 8.00\% \\ \hline
FNN & 2.46 & 4.98 & 5.89\% \\ \hline
FC-LSTM & 2.37 & 4.96 & 5.70\% \\ \hline
STGCN & 2.49 & 5.69 & 5.79\% \\ \hline
DCRNN & 2.07 & 4.74 & 4.90\% \\ \hline
Graph Wevenet & 1.95 & 4.52 & 4.63\% \\ \hline
GMAN & 1.86 & 4.32 & 4.31\% \\ \hline
\multicolumn{4}{|c|}{Training on LA and testing on PEMS-BAY}\\ \hline
TL-DCRNN & 2.13 $\pm$ 1.09 & 5.23 $\pm$ 2.29 & 5.55 $\pm$ 4.34 \\ \hline
\end{tabular}
\vspace{-0.5cm}
\label{tab_comp}
\end{table}


%% file: rel.tex
Recently, graph-convolution-based forecasting models have shown a  significant improvement in traffic forecasting tasks over  classical forecasting approaches such as ARIMA and Kalman filtering, which are not effective in capturing complex spatial and temporal correlations \cite{williams2003modeling, chan2012neural,karlaftis2011statistical, castro2009online}. Cui et al. \cite{cui2018traffic} developed a graph convolutional long short-term memory network. They used the graph convolution operation inside the LSTM cell with regularization approaches. 
Yu et al. \cite{yu2017spatio} integrated graph convolution
and gated temporal convolution in a spatiotemporal convolutional block for traffic forecasting. Li et al. \cite{li2017diffusion} proposed a DCRNN method that models traffic state as a diffusion process on a graph and used it within a GRU cell. All these methods, however, cannot perform forecasting on unseen graphs because they learn location-specific traffic dynamics and require the same highway network for both training and inference.

The prior work on transfer learning for short-term traffic forecasting is sparse. Wang et al. \cite{wang2018cross} proposed an image-based convolutional LSTM network to perform transfer learning for crowd flow prediction from a data-rich city to a data-scarce city. The method first learns a matching function using Pearson correlation to find a  similar source city for each target city. During training of the network the method tries to minimize the hidden representations of the target region and its matched source region inside the loss function. This approach does not incorporate multiple nodes, however, and does not take into account the spatial graph dependency. Recently, Yao et al. \cite{yao2019learning} proposed a metalearning method for traffic volume and water quality prediction. This approach captures knowledge from multiple nodes. It uses an image-based convolutional LSTM network to train on multiple source nodes and uses those trained weights for prediction on the target nodes. The method uses spatial-temporal memory to store representation of diffident regions of the source cities. The regions are found by  $k-$means clustering on the averaged 24-hour patterns of each region, and region-specific weights stored in the memory are utilized for prediction via an attention mechanism. Krishnakumari et al. \cite{krishnakumari2018understanding} developed a method that first clusters the feature vectors obtained from the pretrained image-based convolutional network and then uses the cluster to predict one-step forecast for the similar target location using an ensemble of multiple models such as  multilayer perceptron, random forest, K-nearest neighbor, support vector machine (SVM), and Gaussian process. 
Xu1 et al. \cite{xu2016cross} and Lin \cite{lin2018transfer} conducted preliminary studies for traffic prediction using cross city transfer leaning using SVM and dynamic time warping, respectively. 
Fouladgar et al. \cite{fouladgar2017scalable} proposed a transfer learning method using an image-based convolutional network and LSTM for traffic forecasting in case of congestion. 
None of these methods use graph convolution to model the spatial dependencies, and they cannot be applied directly to  short-term highway forecasting.

\tlDCRNN is inspired by the cluster-GCN \cite{chiang2019cluster} training, where a graph convolution network training is proposed for learning tasks on large graph classification problem. In this approach, each batch for stochastic-gradient-descent-based training uses samples of subgraphs of the original graph. Cluster-GCN is built for node and link classification on graphs, however, and it cannot be used to model graph diffusion and temporal characteristics of the traffic data.

%% file: conc.tex
We developed \tlDCRNN, a graph-partitioning-based transfer learning approach for the diffusion convolution recurrent neural network to forecast short term traffic on a highway network. \tlDCRNN partitions the source highway network into a number of regions and learns the spatiotemporal traffic dynamics as a function of the traffic state and the network connectivity by marginalizing the location-specific patterns.
The trained model from \tlDCRNN is then used to forecast traffic on unseen regions of the highway network. We demonstrated the efficacy of \tlDCRNN using one year of  California traffic data obtained from the Caltrans Performance Measurement System. Moreover, we showed that a model trained with the \tlDCRNN approach can perform transfer learning between LA and SFO regions. 
\tlDCRNN outperformed popular methods used for large-scale traffic forecasting (autoregressive integrated moving average, support vector regression, feed-forward neural network, fully connected LSTM, spatiotemporal graph convolutional network) despite being applied to a region unseen in training, whereas the other methods were both trained and applied on the same region. These results offer strong evidence that practitioners and researchers can begin applying state-of-the-art forecasting methods such as \tlDCRNN to their own regions even in the absence of significant amounts of historical data. Allowing practitioners to apply emerging data-driven methods trained on datasets collected elsewhere is a transformative capability, enabling a wide range of transportation system operations and functions to operate more efficiently and sustainably through improved forecasting at reduced infrastructure development costs.  

Our future work will include (1) developing deployment strategies for Traffic Management Systems, which can vary across the country; (2) transfer learning capability for alternate data sources such as  mobile device data to relieve the cost of installing infrastructure sensors; and (3) metalearning strategies for graph-based transfer learning for highway networks; (4) road network structural implications for extending this approach beyond highway implementations, which may include characterizing how graph constraints are codified in the DCRNN.

\section*{Acknowledgments}
This material is based in part upon work supported by the U.S. Department of Energy, Office of Science, under contract DE-AC02-06CH11357. 
This research used resources of the Argonne 
Leadership Computing Facility, which is a DOE Office of Science User Facility under contract DE-AC02-06CH11357. 
This report and the work described were sponsored by the U.S. Department of Energy (DOE) Vehicle Technologies Office (VTO) under the Big Data Solutions for Mobility Program, an initiative of the Energy Efficient Mobility Systems (EEMS) Program. David Anderson and Prasad Gupte, the DOE Office of Energy Efficiency and Renewable Energy (EERE) managers played important roles in establishing the project concept, advancing implementation, and providing ongoing guidance.

%% file: supplement.tex
\subsection*{Data preprocessing}

We downloaded the dataset from the official PeMS website \cite{pems}. 
PeMS  has been harvesting data from 18,000 sensors, including inductive loops, side-fire radar, and magnetometers. These sensors record the speed and flow of the traffic every 30 seconds; the recorded values are then aggregated at a 5 minutes granularity. Based on PeMS wedsite statistics, due to various types of failures, at any point of time, on average only 69.59\% of the 18,000 sensors are in working condition. The senors report NULL values when they fail and we excluded those failed sensors from the dataset. As a result, we had 11,160 sensors in the final data for the entire year of 2018. Even on the selected 11,160 sensors, there were missing values for several time stamps in the time series data. We found that 0.06\% (698,162 out of 1,173,139,200) data points had missing speed values. We filled the missing data by taking the mean of similar time and day of the week over a period of time. We separated holidays from normal working days. 

The dataset is saved in hdf5 file format. Each column of the hdf5 is the time series data for a senor and the indices are the timestamps in 5 minutes frequency starting from the time 00:00:00, $1st$ January, 2018 to 23.55.00, $31^{st}$ December, 2018. The data used for the experiments can be downloaded at the following link: 


\href{https://bit.ly/2SsGleK}{\color{blue}{https://bit.ly/2SsGleK}}.

To fetch data efficiently for training, we created TFrecord dataset using the hdf5 file. The input pipeline fetches the data for the next batch before finishing the current batch. We use the tf.data API for this purpose. The script used to convert the hdf5 file to TFrecord dataset is given in following link:

\href{https://bit.ly/2SqGtLv}{\color{blue}{https://bit.ly/2SqGtLv}}

Besides the time series data, PeMS  captures spatial information such as the postmile markers for each sensor and latitude and longitude of each postmile markers. To get the latitude and longitude of each sensor, we matched the postmile markers.  The latitude and longitude is found by linear interpolation if an exact match can not be found in the post mile markers. The latitude and longitude are used get the driving distance between the sensors. We used the Open Source Routing Machine (OSRM) \cite{osrm} docker solution to compute the road network distance. 
OSRM takes the latitude and longitude of two sensor IDs and compute the shortest driving distance between them. We limited the OSRM queries only to the 30 Euclidean nearest neighbors, which we precomputed for for sensor ID. The road network distance between the sensors are used to make the adjacency matrix for DCRNN. The computed adjacency matrix for 11,160 sensors is available at the following link:

\href{https://bit.ly/2OS9Fsy}{\color{blue}{https://bit.ly/2OS9Fsy}}.

The training and test datasets are normalized by using a standard scalar method in scikit-learn \cite{scikit-learn}. The normalized feature $Z = \frac{X-\mu}{\sigma}$, where $X$ is the input features $\mu$ is the mean of the training samples, and $\sigma$ is the standard deviation of the training samples. We applied inverse of the transform before computing the MAE on the test data set.

\subsection*{Hardware}
For the experimental evaluation, we used Cooley, a GPU-based cluster at the Argonne Leadership Computing Facility. It has 126 compute nodes, each node consisting of two 2.4 GHz Intel Haswell E5-2620 v3 processors (6 cores per CPU, 12 cores total), one NVIDIA Tesla K80 (two GPUs per node), 384 GB of RAM per node, and 24 GB GPU RAM per node (12 GB per GPU).  The compute nodes are interconnected via an InfiniBand fabric.

\subsection*{Software settings}
We used Python 3.6.0, TensorFlow 1.3.1, NumPy 1.16.3, Pandas 0.19.2, and HDF5 1.8.17. We use Metis 5.1.0 for graph partitioning. Multilevel $k$-way partitioning algorithm is used for the partitioning. This algorithm creates roughly $k$ equal sized partitions. It takes approximately 0.030 seconds to perform 64 partition on a graph of 11,160 nodes.

\subsection*{Hyperparameters}
We used the same hyperparameter values for all the training.  These  hyperparameter values were set as default values for the open-source DCRNN implementation \cite{li2017diffusiongit}. They are batch size: 64; filter type: random walk; maximum diffusion steps: 2; number of RNN layers: 2; number of RNN units per layers: 16;  threshold \texttt{max\_grad\_norm} to clip the gradient norm to avoid exploring gradient problem of RNN \cite{pascanu2013difficulty}: 5; initial learning rate: 0.01 and learning rate decay: 0.1.

\subsection*{Source code}
The code for \tlDCRNN is available at the following URL along with the script to run it:

\href{https://github.com/tanwimallick/TL-DCRNN}{\color{blue}{https://github.com/tanwimallick/TL-DCRNN}}.

\clearpage
\pagestyle{empty} 
\begin{figure*}
\begin{center}
    \framebox{\parbox{6in}{
    The submitted manuscript has been created by UChicago Argonne, LLC, Operator of Argonne National Laboratory (``Argonne''). Argonne, a U.S. Department of Energy Office of Science laboratory, is operated under Contract No. DE-AC02-06CH11357. The U.S. Government retains for itself, and others acting on its behalf, a paid-up nonexclusive, irrevocable worldwide license in said article to reproduce, prepare derivative works, distribute copies to the public, and perform publicly and display publicly, by or on behalf of the Government. The Department of Energy will provide public access to these results of federally sponsored research in accordance with the DOE Public Access Plan. \url{http://energy.gov/downloads/doe-public-access-plan}}}
    \normalsize
\end{center}
\end{figure*}